# Automating Customer Needs Analysis: A Comparative Study of Large Language Models in the Travel Industry

**Simone Barandoni**
Department of Computer Science
University of Pisa
Largo B. Pontecorvo 3, Pisa, Italy
simone.barandoni@phd.unipi.it
0009-0008-4454-3388

**Filippo Chiarello**
Department of DESTEC
University of Pisa
Largo L. Lazzarino, Pisa, Italy
filippo.chiarello@unipi.it
0000-0001-9857-0287

**Lorenzo Cascone**
Department of Computer Science
University of Pisa
Largo B. Pontecorvo 3, Pisa, Italy
l.cascone2@studenti.unipi.it
0009-0001-8707-6360

**Emiliano Marrale**
Department of Computer Science
University of Pisa
Largo B. Pontecorvo 3, Pisa, Italy
e.marrale@studenti.unipi.it
0009-0005-6727-3753

**Salvatore Puccio**
Department of Computer Science
University of Pisa
Largo B. Pontecorvo 3, Pisa, Italy
s.puccio1@studenti.unipi.it
0009-0000-7577-8149

## Abstract

In the rapidly evolving landscape of Natural Language Processing (NLP), Large Language Models (LLMs) have emerged as powerful tools for many tasks, such as extracting valuable insights from vast amounts of textual data. In this study, we conduct a comparative analysis of LLMs for the extraction of travel customer needs from TripAdvisor and Reddit posts. Leveraging a diverse range of models, including both open-source and proprietary ones such as GPT-4 and Gemini, we aim to elucidate their strengths and weaknesses in this specialized domain. Through an evaluation process involving metrics such as BERTScore, ROUGE, and BLEU, we assess the performance of each model in accurately identifying and summarizing customer needs. Our findings highlight the efficacy of open-source LLMs, particularly Mistral 7B, in achieving comparable performance to larger closed models while offering affordability and customization benefits. Additionally, we underscore the importance of considering factors such as model size, resource requirements, and performance metrics when selecting the most suitable LLM for customer needs analysis tasks. Overall, this study contributes valuable insights for businesses seeking to leverage advanced NLP techniques to enhance customer experience and drive operational efficiency in the travel industry.

## 1. Introduction

In the era of vast digital content, understanding customer needs and preferences has become pivotal for businesses across various sectors. The proliferation of social media and online review platforms has provided a wealth of valuable information regarding consumer experiences and sentiments (Vlačić et al., 2021). Extracting and analysing these insights manually, however, can be a daunting and time-consuming task (Zhou et al., 2020). Over the years, the adoption of Artificial Intelligence (AI) and Natural Language Processing (NLP) techniques has facilitated the automation of these processes, especially when working with noisy and unstructured text (Naqvi et al., 2024). However, with the diffusion of Large Language Models (LLMs), this capability can be enhanced even further, promising a new era of automated language understanding and customer needs extraction (Soni, 2023). The utilization of conversational agents powered by LLMs has the potential to reshape the customer needs management area, offering unparalleled efficiency and scalability. By harnessing the power of natural language understanding, these agents can interpret and respond to customer inquiries, identify solutions for new product development, and personalize recommendations (Wu et al., 2023). This paradigm shift might not only enhance user experience, but also optimize operational workflows for businesses, leading to improved customer satisfaction and loyalty.

In recent years, the business sector is experiencing a rapid growth in utilizing conversational agents to enhance the quality and efficiency of the processes (Bavaresco et al., 2020). The selection of the most suitable LLM to build an AI-based chatbot is a critical decision for enterprises, considering both performance and resource requirements. Very large models typically demand substantial computing capabilities, which may pose challenges for some businesses in terms of affordability and maintenance. The current landscape of conversational AI is dominated by proprietary models, such as *GPT-4*, *Gemini*, and *Claude*, with GPT-4 being the state-of-the-art in many Natural Language Processing tasks at the time of the analysis (Hackl et al., 2023), which were performed in March 2024. These models can be utilized for a fee, through APIs. On the other hand, open-source generative AI models, such as LLama 2, are demonstrating good advancements in terms of performance, even if they have yet to reach the level of GPT-4 in certain tasks or specific domains (Ray, 2024). These open LLMs enhance transparency by openly sharing their training data and architectures. Furthermore, considering the possibility of fine-tuning a LLM to enhance its performance within a specific context, open-source models allow users to perform this process on their own infrastructure, providing a level of customization and data security that is not achievable when utilizing a proprietary API (Yang et al., 2023). In the domain of customer management, this is crucial to ensure the utmost protection for any sensitive customer data. There is a vast array of open LLMs available for businesses to leverage in their operations. Therefore, the optimal choice is a non-trivial problem. To the best of our knowledge, no comparisons between LLMs exist in literature regarding the task of needs analysis.

In this paper, we present a comprehensive comparative analysis of Large Language Models for the extraction of travel customer needs from TripAdvisor and Reddit posts. We decided to focus on the travel industry for two reasons: first, because of the great amount of unstructured data that can be easily retrieved about it (e.g., travellers reviews, comments, posts, and requests shared publicly on social media and forums); second, due to the importance of performing a customer needs analysis for enterprises related to the tourism sector, such as hotels, restaurants, transportation companies, travel agencies, etc. These actors may exploit the outcomes of this study to make an informed decision about the selection of the most suitable LLM for building a conversational agent able to support them in identifying customer needs and, consequently, gaining insights about relevant design requirements (Büyüközkan et al., 2007). By employing a process of data collection, labelling, and computational analysis, we seek to elucidate which LLMs excel in this specialized domain. The models considered in the analysis are the following: *Llama 2 7B, Llama 2 13B, Phi-2, Mistral, Mixtral, Gemma, Gemini 1.0, GPT-3.5*, and *GPT-4*. This selection allowed us to explore not only the performance differences between open-source (Llama 2, Phi-2, Mistral, Mixtral, Gemma) and proprietary models (GPT-3.5, GPT-4, Gemini) but also the impact of model size and type of prompting on the quality of results.

By needs we intend both essential requirements and preferences that a service or a place should fulfil to satisfy the customer. In the data we collected (presented in Section 3.2), users share their requests about a certain place, to receive feedback and suggestions, usually expressing one or more needs. Even if the needs are not explicitly written in the text, they can be extrapolated from it analysing its keywords. Therefore, we first used a thematic analysis approach to manually annotate the dataset: we read the collected posts and extracted recurring themes to identify customer needs. Then, we employed the mentioned LLMs to perform the same task and compared their outcomes with the manual labelling. Figure 1 presents an example of post and corresponding customer needs identified through thematic analysis.

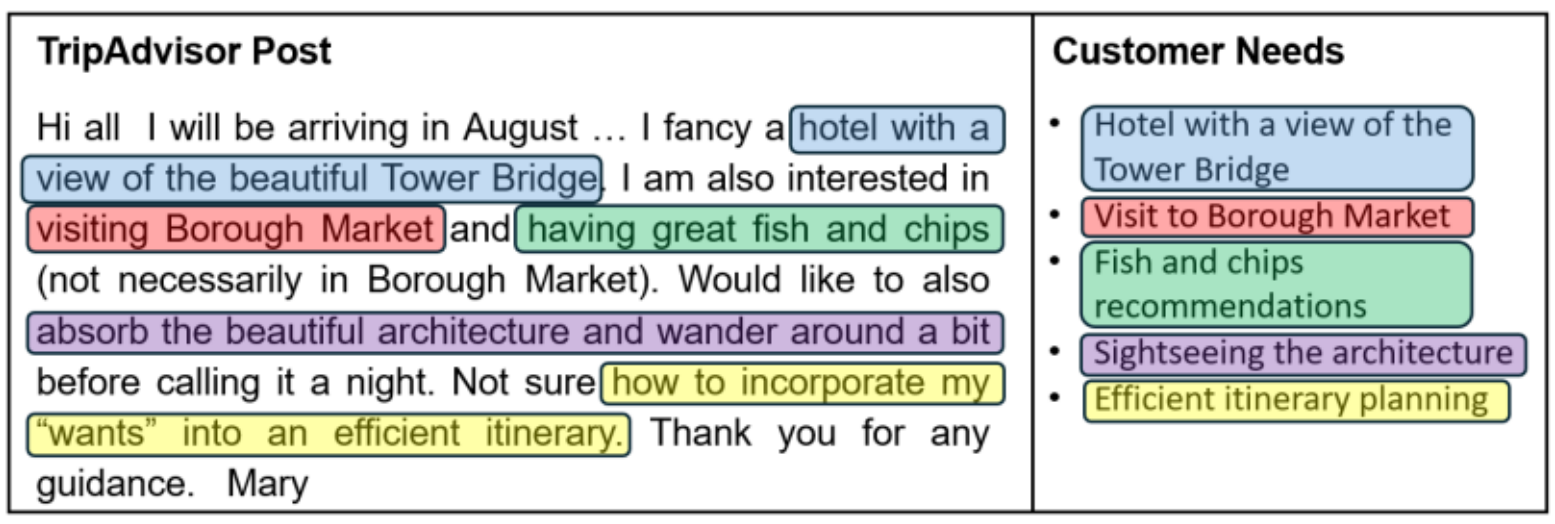

Figure 1. Example of a TripAdvisor post and related customer needs manually extracted through thematic analysis. Some of the needs, such as *Hotel with a view of the Tower Bridge* are explicit in the original text. Others, such as *Efficient itinerary planning*, are more implicit.

## 2. Related Works

The related works of this study can be organized in two areas: the first, addressed in Section 2.1, revolves around machine learning for customer needs analysis, discussing the data sources and the methodologies employed; the second, in Section 2.2, concerns the utilization of Large Language Models in the domain of Customer Experience Management.

### 2.1 Utilizing User-Generated Content and Machine Learning for customer needs analysis

Researchers and practitioners have exploited various sources of data and developed various methodologies to automatically analyse user needs and to provide support in business strategy (Timoshenko and Hauser, 2019).

User-Generated Content (UGC) is the main source of data utilised for these purposes (Kilroy et al., 2022). UGC refers to any form of content, such as text, images, videos, or reviews, which is created and shared online by users of a platform or website (Spada et al., 2023). This kind of data offers unfiltered insights into the authentic experiences and opinions of customers and individuals, without being subject to modification or censorship by traditional media outlets (Krumm et al., 2008). Social media posts and product reviews are the most utilised UGC for customer needs analysis (Kilroy et al., 2022). Social media such as X (formerly Twitter), Facebook, Instagram, often contains irrelevant posts, unlike product reviews which are more focused. However, for researching future needs, social media may be more suitable as their users tend to discuss emerging trends (Kilroy et al., 2022). In this work, we utilised the TripAdvisor forums and Reddit to collect textual documents. The first is a question-and-answer social media specifically focused on travels, therefore we considered it appropriate for our purposes (i.e., testing LLMs to predict travel customer needs). In order to obtain a broader perspective, rather than relying on a single source, we also employed Reddit. This social media is based on communities of users (subreddits) related to specific topics or domains, and has been employed extensively in many research domains, tourism included (Hao et al., 2024; Medvedev et al., 2019). r/Travel is a subreddit where users request travel trips, experiences, and advice, similarly to TripAdvisor's Forums. In section 3.2 we provide a better description of the TripAdvisor forums and the r/Travel subreddit and of the reasons for their selection as a source of UGC.

Regarding the methodological aspects, these usually depend on the kind of data used and on the specific needs to be identified. Researchers have exploited both traditional NLP techniques based on lexicons and rules (Ding et al., 2008; Abrahams et al., 2015; Spada et al., 2023) and machine learning approaches: Zhou et al. (2020) employed Latent Dirichlet Allocation (LDA), Sentiment Analysis, and a Kano model to analyse customer needs for product ecosystems; Timoshenko and Hauser (2019) trained a convolutional neural network (CNN) and developed a machine-human hybrid approach to extract customer needs from UGC; Haque et al. (2018), applied Sentiment Analysis to a set of Amazon reviews; Bigi et al. (2022) collected a series of questions and answers from TripAdvisor forums and analysed them through a Bayesian machine-learning technique to investigate the relevance of food and drink products in attracting visitors. In recent times, it has become evident that the most effective models across various NLP tasks and domains are those rooted in deep learning methodologies. Luo et al. (2021) corroborated this by highlighting that deep learning models outperform other machine learning algorithms in customer needs analysis as well, in a study which involved analysing Yelp's restaurant reviews. This trend extends to the latest advancements in Generative AI and LLMs. For this reason, we focused our study on Large Language Models.

### 2.2 Large Language Models and Customer Experience Management

The recent diffusion of conversational agents based on LLMs, particularly *ChatGPT*, has fuelled the scientific research about applications and potential impact of these technologies across various domains, such as medicine and healthcare (Gilson et al., 2023; Li et al., 2024), scientific research (Burger et al., 2023), education (Haman and Školník, 2023), (Mhlanga, 2023; Lo, 2023), and many others. Customer experience and feedback management are two of the areas which might be revolutionized through an implementation of such AI-powered systems. ChatGPT and similar conversational AI might be utilized to provide improved consumer assistance, by responding to queries in a better way than traditional chatbots, recommending products, and to analyse customer needs (Haleem et al., 2022). Researchers have explored various aspects of employing these technologies to better understand and fulfil customer requirements. Kim et al. (2023) demonstrated the positive impact of implementing ChatGPT into recommender systems in terms of consumer satisfaction. Kumar et al. (2023) argued that, despite the current limitations, such as difficulties in understanding non-standard languages, ChatGPT will become an important asset for retailers, as a tool to support the processes of consumer needs analysis. Trichopoulos et al. (2023) proposed a fine-tuned version of GPT-4 to be used as a recommender system for museums. Soni (2023) explored the use of LLMs in customer lifecycle management and observed their positive impact on improving and optimizing lead identification, audience segmentation, marketing strategies, sales support, and post-purchase engagement. Considering the travel domain, the findings of Mo et al. (2023) suggest that LLMs have the potential to be effective tools for predicting travel behaviour, offering competitive performance and the ability to provide explanations for their predictions. Still, their implementation in the travel sector presents unique challenges and limitations. One significant issue is in the complexity and variability of travel-related data, which can be highly dynamic and context-dependent. For instance, LLMs may struggle to accurately interpret user preferences in travel scenarios due to the cultural, linguistic, and regional differences that influence travel decisions. Additionally, the real-time nature of travel planning requires LLMs to process and integrate different kinds of rapidly changing information, such as flight availability, weather conditions, and geopolitical events. This can be challenging, given the models' dependency on the training or fine-tuning processes. Furthermore, in this sector, data is often sensitive and holds significant commercial value (Song and Liu, 2017). This makes it more difficult to collect and employ the right data for developing accurate and reliable models, as the process also requires careful handling to ensure privacy, security, and compliance with legal regulations.

## 3. Methodology

The methodology of this study encompasses a series of phases, summarized in Figure 2. We started by selecting the LLMs to be assessed and collecting the data for the analysis (a corpus of TripAdvisor posts). Then, we manually labelled the data, identifying the user needs expressed in the texts. The objective was using the models to automatically extract the needs from the documents and then compare the outcomes with the manual labelling to measure their performances. We designed an initial prompt and deployed each LLM performing a prompt optimization (a process aimed to adapt the instructions provided to the model to its characteristics, to improve the quality of the responses), executing the extraction of the needs, and cleaning the results. Finally, we evaluated the performance of each model by comparing the predicted customer needs with those identified in the manual labelling.

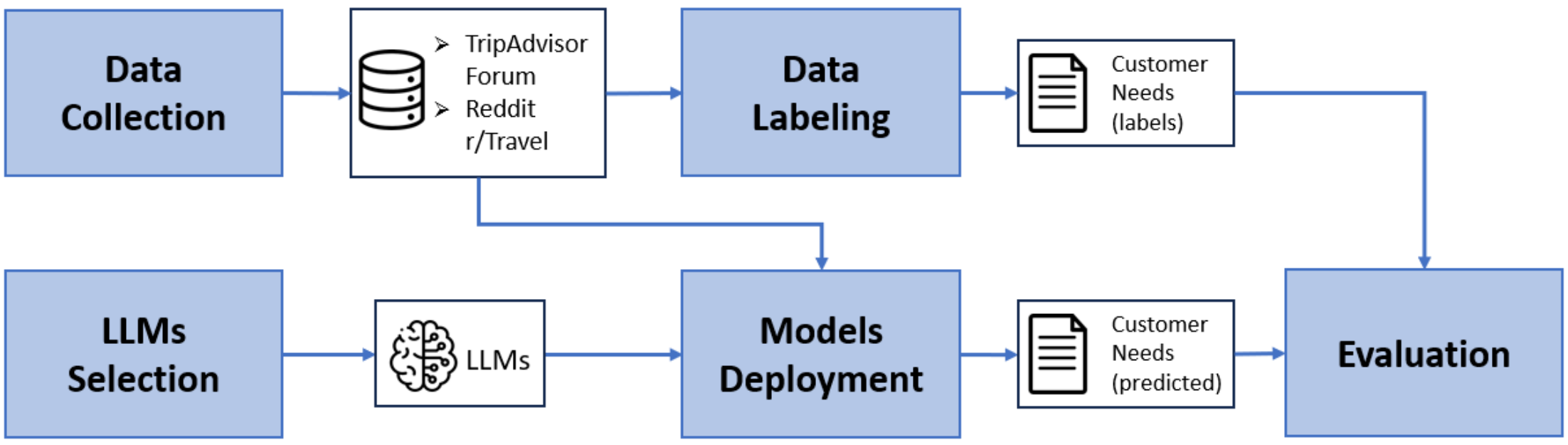

Figure 2. Workflow of the methodology employed to compare the Large Language Models for the task of travel customer needs extraction from TripAdvisor and Reddit posts.

### 3.1 LLMs Selection

Table 1 demonstrates the models we employed in this study. We selected a series of open-source LLMs, except for Gemini 1.0 Pro, GPT-3.5 and GPT-4.

| **Model** | **Parameters** | **Proprietary** | **Reference** |
|---|---|---|---|
| Gemini 1.0 Pro | Not available | Google | Team et al., 2023 |
| Gemma 7B | 7 billion | Google (open-source) | Team et al., 2024 |
| GPT-3.5 | 175 billion | OpenAI | Brown et al., 2020 |
| GPT-4 | Not available | OpenAI | Achiam et al., 2023 |
| Llama 2 7B Chat | 7 billion | Meta (open-source) | Touvron et al., 2023 |
| Llama 2 13B Chat | 13 billion | Meta (open-source) | Touvron et al., 2023 |
| Mistral 7B Instruct | 7 billion | Mistral AI (open-source) | Jiang et al., 2023 |
| Mixtral 8x7B* | 46.7 billion | Mistral AI (open-source) | Jiang et al., 2024 |
| Phi-2 | 2.7 billion | Microsoft (open-source) | Javaheripi et al., 2023 |

Table 1. List of the LLMs utilized in the analysis. For each model we present the version, the number of parameters, the enterprise or the institute which released it, and the reference. *Due to the large computational requirements of Mixtral 8x7B, we were forced to employ its smallest version (*2 Bit quantization*).

Gemini 1.0 is a multimodal model property of Google, available in three versions: Ultra, Pro, and Nano. GPT-3.5 and GPT-4 are the two LLMs published by OpenAI. The first is publicly available for free through ChatGPT, while the second requires a paid subscription. GPT-3.5 has 175 billion parameters (Brown et al., 2020), while for GPT-4 and Gemini 1.0 this information is not available. Llama 2 is a family of pretrained open-source Large Language Models, ranging from 7 to 70 billion parameters (Touvron et al., 2023), released by Meta. This family comprehends a series of models which are optimized for dialogues, the Llama 2-Chat ones. We chose to use the two smaller models, i.e. Llama 2-Chat 7B and Llama 2-Chat 13B, due to the very large computational resources required to deploy bigger models. We assumed that this is a limitation that small and medium enterprises might face, since not all the organizations can rely on a powerful infrastructure of computer systems, able to run the biggest LLMs. Gemma is a family of open-source LLMs released by Google, much smaller than Gemini (Team et al., 2023). It can be utilized in the two variants of two billion and seven billion parameters (Team et al., 2024). We selected Gemma 7B for our analysis. Phi-2 is an open LLM built by Microsoft Research. It is a smaller model, with its 2.7 billion parameters (Javaheripi et al., 2023). Mistral 7B and Mixtral 8x7B are two open models released by Mistral AI. The first is a seven billion parameters LLM (Jiang et al., 2023). The second, considered the most powerful by the developers, has the same architecture, but it is also based on layers "composed of 8 feedforward blocks (i.e. experts). For every token, at each layer, a router network selects two experts to process the current state and combine their outputs." (Jiang et al., 2024).

### 3.2 Data Collection

As mentioned in Section 2.1, we decided to employ the TripAdvisor Forum and Reddit's r/Travel as sources of data. The first is an online platform hosted by TripAdvisor, the most popular website about travelling, according to Xiang and Gretzel (2010). It serves as a community-driven space where travellers can seek and share information, advice, and experiences related to various destinations, accommodations, attractions, and transportation options worldwide. The platform provides a series of destination-specific forums: each of them rely on contributions from users who share their travel experiences, ask questions, and provide answers to queries posted by other users regarding specific cities, regions, and countries around the world. The content shared by the users covers a wide range of topics related to travels, including accommodation reviews, restaurant recommendations, transportation inquiries, and general travel tips. Therefore, it can be utilised for analysis purposes to gain valuable insights into travel trends and consumer preferences (Gal-Tzur et al., 2018). Due to the structure of the website, which is organized in many independent forums, we had to select which one to scrape. We collected the 110 most recently published posts from three different forums of the platform, London, Tokyo, and New York, for a total of 330 documents. We executed this web-scraping operation on 22/12/2023. The collected corpus of texts has a vocabulary size of 3,668 words, and the posts are composed of 101 words, on average.

r/Travel is a subreddit community where users request and exchange information about travels. We chose Reddit because, similarly to TripAdvisor Forum, it encourages question-answering and open discussions, helping users plan trips and share insights and leading to more context-specific conversations than platforms centred around social networking, like Facebook. Additionally, Reddit's anonymity fosters more genuine and unbiased conversations, providing a valuable resource for understanding the experiences of online communities (Hao et al., 2024). We employed a dataset of Reddit posts available on Kaggle[1], which has been collected from the r/Travel subreddit in December 2022. It contains 1,899 posts. We randomly selected 300 of them, to be used in the analysis. These posts are slightly longer than the TripAdvisor ones (on average, 121 words), and the corpus presents a vocabulary size of 4,297 words.

Reddit's posts, while slightly longer than TripAdvisor's, share a similar structure and content. In both platforms, users tend to follow the typical format of a forum discussion, asking questions or recommendations to the other community members. Therefore, we combined the two datasets in a single corpus, for a total of 630 posts. We randomly divided this dataset into a prompt-optimisation set (30 posts) and test set (600 posts). In this study, we used the 30 prompt-optimisation posts for the prompting phases (Section 3.4 will provide the details), and the 600 test posts to interrogate the LLMs and evaluate their answers. As explained in the next section, we manually annotated all of them. The prompt-optimisation set was very small because we did not aim at training or fine tuning the models. We took this decision to ensure that the evaluation of the models remained consistent across the board. Our objective was to assess the raw capabilities of each model without any adjustments or enhancements through fine-tuning. However, we acknowledge that fine-tuning these models could potentially improve their performance, especially in specialized domains like customer needs analysis. We utilised the prompt-optimisation set only in the Prompt Optimisation phase, where a small number of labelled documents was necessary, as explained in section 3.4.1.

## 3.3 Data Labelling

Three authors performed the manual labelling phase separately, by reading all the textual documents collected (630 posts) and identifying the customer needs expressed in each text.
As explained in the Introduction, the annotators performed a thematic analysis on the posts, reading through the text, highlighting relevant keywords and themes, and distilling these into customer needs (Figure 1 provides an example of this process).
The following were the guidelines followed by the three annotators:

- Read the posts.
- Highlight relevant themes, by identifying keywords and expressions related to the travelling preferences, desires, and requests of the user.
- Group together recurrent themes: e.g., use the word *accommodation* instead of all the specific keywords, such as *hotels, apartments*, or *itinerary* instead of *path, route*.
- Summarise the user needs using as few words as possible and without providing too specific information. Consider the location or the period requested (if present) but avoid further details. For example, *"New Year's Eve parties in London"* is allowed, while *"New Year's Eve parties on a rooftop in London starting at 11 p.m."* is too specific.

[1] Reddit: /r/travel: www.kaggle.com/datasets/thedevastator/uncovering-travel-experiences-desires-and-opinio

We decided to measure the inter-rater reliability (IRR) using Fleiss' kappa coefficient (Fleiss et al., 1971), to quantify the level of agreement among the three raters and ensure the consistency of the manual labelling process. The following is the formula of the Fleiss' kappa:

$$K = \frac{P_o - P_e}{1 - P_e}$$

Po is the average observed agreement. In our study, this corresponds to the mean of the proportion of observed agreements calculated for each of the labelled posts. Pe is the expected agreement (i.e., if the rates would have performed the labelling through random assignment). For each post, the proportion of observed agreement between the three authors was calculated by counting the number of matches between the needs they identified and dividing by the number of total possible agreement. However, it may have happened that different raters would write the same concept with different words. Therefore, to compare the three different annotations and measure the IRR, we calculated the semantic similarity between the three sets of entities extracted in each document through text embeddings (by using the pretrained embedding model Roberta Large; Liu et al., 2019) and cosine similarity. Then, a fourth author has examined all the pairs of needs extracted by different authors which maximised the similarity and determined whether they were corresponding to a match (i.e., the two entities were expressing the same concept). These were considered to count the observed agreements between the three raters and calculate the average observed agreement (Po), obtaining 0.81. Regarding the expected agreement (Pe), we reached the conclusion that this should be set equal to 0: for this kind of labelling, the raters did not assign fixed categories to the data. They extracted the needs from the text by reading and identifying relevant keywords, a task that could not involve a random assignment. Therefore, the probability that performing this randomly would result in an agreement is zero. For this reason, the Fleiss' Kappa is equal to Po (0.81). This high result indicates a substantial level of agreement among the raters; therefore, we considered the manual labelling as consistent and reliable. We identified a total of 1071 needs (691 unique ones), and 1.71 needs per post. The final labels were composed of a unified version of the three manual labelling done by the raters: the fourth author collapsed needs which expressed the same concept through different words (and constituted the matches identified through the semantic similarity) into a unique form, and left all the others (the unmatched ones, which were extracted by only one rater) as they were. We included the unmatched ones as well, considering the high level of agreement between the three manual labelling.

### 3.4 Models Deployment

To deploy the selected LLMs we relied on a machine with the following hardware specifications:

- Chip Apple M2 Max with 30 core GPU and 12 core CPU.
- 32 GB of unified ram.
- 1TB SSD.

Furthermore, we utilized GPT-4, GPT-3.5, and Gemini through OpenAI's and Google's API, respectively. We designed an initial prompt to be provided to the models. Prompting refers to the technique of describing specific instructions to a language model to guide its generation of text or response to a given input. It allows teaching LMs to perform new tasks (Kojima et al., 2022). In zero-shot prompting, the prompt is provided without any example. In few-shot prompting, the instructions also include a small number of examples of the desired output. This typically enhances the performance of a model (Brown et al., 2020). Figure 3 demonstrates the initial few-shot prompt created: it presents a series of instructions on the task and three examples derived from the prompt-optimisation set. We chose the three examples randomly but avoiding duplicated needs among them.

```
"""You are a customer service expert. I am going to provide
you a TripAdvisor post about travels, you should identify the
needs that the user expressed in the text. Needs should be
expressed as short sentences. The answer should be generated
as a list of needs separated by a comma. Avoid any unnecessary
text or comment. Here are some examples:
POST: I've come across a Cyber Sale announced by the Sonesta
group (used to be Affinia) for their three hotels in NYC. The
sale goes through April 2024 and looks like the rates are
refundable but you must stay at least 2 nights. Cheapest dates
are January and February as usual but over the Thanksgiving
week / few days it's not too bad either. Any other good offers
at other hotels are welcome to be posted in this thread as
well.
ANSWER: Affordable accomodations, Accomodations advices
POST: We are here in NYC; looking for a last minute show to
see on or off broadway on Thanksgiving day. Any
recommendations sites to purchase?
ANSWER: Broadway musical tickets retrieving
POST: I am traveling from New York to Dc this Sunday. Amtrak
tickets are really expensive due to the holiday and waiting
last minute. Please recommend any discounts or other safe
options for travel. Thank you.
ANSWER: Cheap airline travelling tips and suggestions"""
```

Figure 3. Initial few-shot prompt designed for the task of travel customer needs extraction. The instructions are highlighted in bold, while the rest of the text corresponds to the three examples.

Chain-of-Thought (CoT) is a further prompting technique which involves generating intermediate reasoning steps. It has been demonstrated that generating a chain of thought "significantly improves the ability of large language models to perform complex reasoning" (Wei et al., 2022). In this study, we decided to assess two forms of prompting: standard few-shot and optimized Chain-of-Thought. For the first, we used the initial prompt (Figure 3) for all the models: we iterated over the three hundred posts of the training set to let the LLMs extract the needs. All the models were deployed setting the *temperature* parameter to 0.5. For the latter, we also performed a prompt optimization, which means adapting the prompt to each LLM to improve the quality of the generated answers. We did not perform the optimization step on the initial few-shot prompt to be able to measure the reliability of this phase by comparing the results obtained through an optimized prompt with those obtained through a non-optimized one.

### 3.4.1 Prompt Optimization

Optimizing the instructions provided to a specific LLM typically comprehends designing a series of prompts through manual trial and error (Khattab et al., 2023). This makes the prompting delicate and unscalable, especially when working with more than one LLM at once: a prompt which has been manually optimized for a model might not be adapted for other LLMs. Therefore, we decided to employ a more systematic methodology to optimize the CoT prompting. We relied on DSPy (Khattab et al., 2023), a programming model designed to build and optimize LM pipelines by abstracting them as text transformation graphs and automating the optimization process through parameterized declarative modules and teleprompters. Its employment allowed us to reduce the reliance on manually crafted prompt templates. DSPy comes with a compiler that optimizes a model's pipeline to maximize a given metric. More specifically, we used the *BootstrapFewShotWithRandomSearch* DSPy optimizer, with *max_bootstrapped_demos*=3, *max_labeled_demos*=3, *num_candidate_programs*=10, *num_threads*=8 as parameters, and the *ChainOfThought* module. For each LLM employed, we provided the compiler with the initial few-shot prompt, the thirty labelled posts of the prompt-optimisation set, and BERTScore as validation metric (section 3.5 provides an explanation of this metric). The DSPy optimiser used the 30 examples from the prompt-optimisation set to self-generate reasonings and to craft a CoT prompt that would optimise the output over the given metric. Then, the optimised prompts were used to interrogate all the models to extract the needs from the posts of the test set. Figure 4 demonstrates four examples of prompt-optimisation set instances for which DSPy automatically generated reasonings, to be used to optimise the CoT prompting. These, in particular, are part of the final CoT prompt optimised for Mistral.

**Question:** My family is going to be on the Isle of Skye between Christmas and NYE. A couple of questions: 1) I have found AirBnBs all over the island and wonder if we're better off paying for the more expensive place in Portree and using it as a home base to explore or paying less for a nice waterfront cottage somewhere else? Is it okay to be anywhere on the island this time of year? Is there a preferable location for first time visitors? 2) I wonder if anyone knows what sorts of things are open this time of year. Seems like a lot shuts down for the winter. We still want to go for the natural beauty but wonder if any of the sites will be accessible (like the castle at Dunvegan? Any boat tours? Thanks for any assistance! Cheers Nate
**Reasoning:** Let's think step by step in order to produce the answer. We need to consider the location, budget and what is open during Christmas time on Isle of Skye. The user needs information about a good accommodation and would like to know the opening hours of interesting places.
**Answer:** Accommodation advices, Opening hours information.

**Question:** Hey fellow travel enthusiasts! 👋 So here's the scoop - my [...] hear your recommendations! 🗽🎄✈️
**Reasoning:** Let's think step by step in order to produce the answer. We need to consider the time of day, budget, transportation options and must-see places/restaurants. We also need to take into account the weather conditions.
**Answer:** Must-Visit Places/Restaurants, Transportation advice, Budget considerations.

**Question:** Where could I deposit luggage nearby Penn Station for 12 hours? Leaving it there in the morning and picking it up in the evening before heading to JFK Airport. Is Bounce app a good option?
**Reasoning:** Let's think step by step in order to produce the answer. The user needs information about a place to deposit luggage for 12 hours near Penn Station, preferably one that is open early in the morning and late in the evening. We also want to make sure it is safe and secure. Bounce app is a popular option for luggage storage but we need to check if it is available at Penn Station and if it meets our requirements.
**Answer:** Luggage storage advices.

**Question:** Hi I'm looking for some recommendations of where we can take our son and now finance to celebrate their engagement. Looking to spend about £60-70 each for two courses without wine. Somewhere romantic and memorable. They live in Kidbrooke but we will be seeing Guys and Dolls in the evening so anywhere in central London would be great. This is for lunch. Thank you in advance. This is a February visit.
**Reasoning:** Let's think step by step in order to produce the answer. The user is looking for a romantic and memorable restaurant that serves two courses for £60-70 each without wine, preferably in central London as they will be seeing Guys and Dolls in the evening. We also need to consider the budget and location.
**Answer:** Restaurant recommendations.

Figure 4. Examples of posts from the prompt-optimisation set employed by DSPy to generate reasonings for the CoT prompting optimisation.

### 3.4.2 Results Cleaning

We screened the output produced by each model with both prompting approaches to check whether they generated a clean answer (containing the list of identified customer needs only, following the instruction provided in the prompt). The Gemma, Llama 2, Mixtral, and Phi-2 models happened to add additional text in some of the responses, usually to introduce the list of extracted needs (e.g., "*Sure, these are the needs: …*"). We eliminated all the additional texts through regular expressions to consider only the needs in the Evaluation phase. Since each model generated different types of superfluous sentences, we did not employ a unique methodology to clean all the LLMs' outputs, but we adapted the regular expressions to each specific case. Furthermore, we deleted symbols and numbers from the results' texts.

## 3.5 Evaluation

This section describes the evaluation process and the metrics chosen to measure the performance of each model. We aimed at comparing the texts generated by the LLMs with the manual labelling. We performed this comparison through the following metrics: BERTScore, ROUGE, and BLEU.

BERTScore is an automatic metric for evaluating text generation which computes Precision, Recall, and F1 score by leveraging pretrained text embeddings and cosine similarity (Zhang et al., 2019). Precision expresses the ratio of matching words in the candidates (in our case, the generated texts of a LLM) to those in the reference (the manual labelling); Recall calculates the ratio of matching words in the reference to those in the candidate; F1 score, a harmonic mean of Precision and Recall, offers a balanced appraisal of a model's effectiveness (Sasaki, 2007). We utilized the *deberta-xlarge-mnli* pretrained embedding model. At the time of the analysis, this was the best model for running BERTScore in terms of correlation with human evaluation, according to the authors of BERTScore (Wu, 2020). *DeBerta* is a model architecture proposed to enhance existing pre-trained neural language models, particularly *BERT* (Devlin et al., 2018) and *RoBERTa* (Liu et al., 2019), by introducing two novel methodologies, *Disentangled attention mechanism* and *Enhanced mask decoder* (He et al., 2020).

ROUGE is a well-established evaluation metric, designed for text summarization tasks. It counts "the number of overlapping units such as n-gram, word sequences, and word pairs between the computer-generated summary to be evaluated and the ideal summaries created by humans" (Lin, 2004). In this study, the task performed by LLMs is not precisely a summarization, nor a machine translation, so we first rely on the BERTScore metrics. However, in some cases we can also interpret it also as summarization: it can happen that users did not explicitly write a need in a TripAdvisor post, but expressed it through longer sentences, that LLMs should be able to summarize in a few words. Therefore, we decided to calculate the ROUGE metric as well. More specifically, we measured the F1 score of the number of overlapping unigrams (ROUGE-1) and of the longest common subsequence (ROUGE-L). Furthermore, we also considered the BLEU metric, which is usually utilized in machine translation to measure the quality of generated texts by comparing it with the reference ones through "a weighted geometric mean of n-gram precision scores" (Celikyilmaz et al., 2020). This holistic approach ensures a comprehensive evaluation of the models' performance.

## 4. Results

This section describes the results of the work, presenting both the statistics of the extraction performed through the LLMs and the evaluation of their output. Table 2 demonstrates the number of needs extracted by the models and the manual labelling. It is possible to observe that almost all the LLMs identified more entities than the human annotators. The models whose statistical results are more in line with the manual labelling are Gemini (through Chain-of-Thought), GPT-3.5 (few-shot), GPT-4, and Mistral. With the results of the Evaluation step, we can observe whether this statistical alignment is also reflected in the similarity between the manual labelling and the models' output, helping to determine if the identified needs are consistent not only in quantity but also in qualitative accuracy.

| Approach | Prompt | Extracted Needs |
|---|---|---|
| Manual Labelling | - | 1,078 |
| Gemini 1.0 Pro | Chain-of-Thought | 1,144 |
| | Few-shot | 1,501 |
| Gemma 7B | Chain-of-Thought | 2,203 |
| | Few-shot | 1,756 |
| GPT-3.5 | Chain-of-Thought | 1,479 |
| | Few-shot | 1,064 |
| GPT-4 | Chain-of-Thought | 1,266 |
| | Few-shot | 1,291 |
| Llama 2 7B | Chain-of-Thought | 1,496 |
| | Few-shot | 1,239 |
| Llama 2 13B | Chain-of-Thought | 1,787 |
| | Few-shot | 1,155 |
| Mistral 7B | Chain-of-Thought | 1,017 |
| | Few-shot | 947 |
| Mixtral 8x7B | Chain-of-Thought | 2,389 |
| | Few-shot | 1,722 |
| Phi-2 | Chain-of-Thought | 1,590 |
| | Few-shot | 1,235 |

Table 2. Statistical results of the needs extraction performed by the LLMs. For each approach, the type of prompt and the number of identified entities are shown. The statistics of the manual labelling are included as well, providing a baseline to compare the models' outputs against human annotations.

As presented in section 3.5, we evaluated the output of each model (obtained both through CoT and few-shot prompting) by comparing the predicted customer needs with the ones manually identified in the Data Labelling. To perform this comparison, we calculated the average BERTScore, Rouge-1, Rouge-L, and BLEU. Table 2 presents the results of the evaluation, showing the calculated metrics.

| LLM | Prompt | BERTScore | | | Rouge-1 | Rouge-L | BLEU |
|---|---|---|---|---|---|---|---|
| | | Precision | Recall | F1 | F1 | F1 | |
| GPT-4 | Chain-of-Thought | **0.704** | **0.683** | **0.693** | **0.471** | **0.454** | **0.513** |
| Mistral | Chain-of-Thought | **0.634** | 0.653 | **0.643** | 0.341 | 0.331 | 0.326 |
| Mistral | Few-shot | 0.618 | **0.659** | 0.638 | **0.364** | **0.350** | **0.348** |
| Gemini | Chain-of-Thought | 0.613 | 0.612 | 0.613 | 0.312 | 0.292 | 0.310 |
| GPT-3.5 | Few-shot | 0.632 | 0.590 | 0.611 | 0.284 | 0.244 | 0.362 |
| GPT-4 | Few-shot | 0.633 | 0.544 | 0.585 | 0.252 | 0.204 | 0.360 |
| GPT-3.5 | Chain-of-Thought | 0.616 | 0.544 | 0.578 | 0.243 | 0.214 | 0.367 |
| Gemini | Few-shot | 0.567 | 0.520 | 0.542 | 0.184 | 0.159 | 0.286 |
| Llama 2 7b | Chain-of-Thought | 0.602 | 0.471 | 0.528 | 0.156 | 0.136 | 0.276 |
| Phi-2 3b | Chain-of-Thought | 0.556 | 0.461 | 0.504 | 0.147 | 0.135 | 0.192 |
| Mixtral | Chain-of-Thought | 0.571 | 0.446 | 0.501 | 0.115 | 0.104 | 0.255 |
| Llama 2 13b | Chain-of-Thought | 0.608 | 0.422 | 0.499 | 0.124 | 0.106 | 0.285 |
| LLama 2 13b | Few-shot | 0.592 | 0.424 | 0.495 | 0.122 | 0.103 | 0.279 |
| LLama 2 7b | Few-shot | 0.576 | 0.425 | 0.489 | 0.136 | 0.116 | 0.286 |
| Mixtral | Few-shot | 0.541 | 0.422 | 0.474 | 0.079 | 0.068 | 0.215 |
| Phi-2 3b | Few-shot | 0.546 | 0.401 | 0.462 | 0.093 | 0.083 | 0.215 |
| Gemma7b | Few-shot | 0.544 | 0.352 | 0.427 | 0.072 | 0.061 | 0.231 |
| Gemma7b | Chain-of-Thought | 0.546 | 0.312 | 0.397 | 0.053 | 0.040 | 0.204 |

Table 3. Results of the comparative analysis. For each LLM, the table presents the type of prompting employed and the scores obtained through the different metrics. The highest achieved values with proprietary and open-source models are highlighted in bold.

GPT-4, executed through optimized Chain-of-Thought prompting, has achieved the highest score in all the metrics employed. Mistral 7B is the second-best model, both with optimized CoT and non-optimized few-shot prompting: it outperformed all the other LLMs, Gemini and GPT-4 (standard few-shot) included. Despite being much smaller than the closed models assessed (GPT-4, GPT-3.5, and Gemini 1.0) in terms of numbers of parameters, it achieved a remarkably high quality of the responses. Surprisingly, GPT-3.5 achieved slightly better results through non-optimized than through optimized prompting. Except for Mistral, all the open-source models obtained lower scores than the closed ones, especially through non optimized few-shot prompting. Gemma7b was the worst model in terms of accuracy.

In terms of selecting the best model to perform travel customer analysis, we concluded that the best solution is Mistral 7B. It provided an extremely high quality of responses, the same of GPT-4, and its small size (seven billion parameters) allows it to be deployed on affordable machines. Furthermore, being open-source, it enables an elevated level of customization and security, as explained in the Introduction. Among the other open-source models, we observed satisfactory results also in Llama 2 7B and Phi-2 3B. With only three billion parameters, Phi-2 can be run on even smaller computer infrastructures. Therefore, it might be taken into consideration by practitioners and researchers with limited resources. Finally, we believe that it is possible to achieve much better scores with Llama 2 13B and Mixtral7Bx8: being larger than Llama 2 7B and Mistral, respectively, there is the possibility that their performance has been negatively influenced by our limited hardware.

### 4.1 Discussion on the evaluation metrics employed

We tried to interpret the results achieved through BERTScore by reading a sample of the output of each model, to estimate the quality of the answers also through human evaluation. We noticed that the needs generated by the models which achieved a F1 (BERTScore) of more than 0.60 were of extremely high quality: they corresponded to those identified in the manual labelling, usually even adding more specific information. Table 3 demonstrates three examples of this situation: the needs extracted by GPT-4 are consistent with the label's ones, but they provide extra content. For instance, in the third example of Table 3, we manually identified the need *Bar recommendations*; GPT-4 added more specific information about the type of bar, generating *Pop-up bar recommendations*.

| Post | Label Needs (Manual Labelling) | Generated Needs (GPT-4 Chain-of-Thought) |
|---|---|---|
| Is the journey back after the Staten Island Ferry just getting you back to Times Square. We are thinking of missing out on the Ferry and finishing the rest of the day in Lower Manhatten would we be missing much by doing this? | Itinerary planning advices | Travel itinerary advice, Information about Staten Island Ferry, Recommendations for activities in Lower Manhattan |
| Hi! Visiting in 2.5 weeks with hubby and 18 year old daughter. We're going ice skating in Central Park on our final night and fancy going for a Mexican afterwards - any recommendations? We're flexible on budget we will have spent the day out and won't be dressed up so nowhere too fancy! | Restaurants advices | Mexican restaurant recommendations, Casual dining options |
| We are 6 crazy older women from Texas and wanting to try one of these pop-up bars out in December. Miracle on 9th or Frosty's? Thank you! | Bar recommendations | Pop-up bar recommendations |

Table 4. Examples of needs extracted by GPT-4 through Chain-of-Thought prompting and manually identified in three TripAdvisor posts.

The additional information provided by the LLMs are useful in terms of needs analysis, but could lower the scores of the evaluation metrics, because the generated sentences contain more words than the reference ones: this can decrease their semantic similarity. Therefore, we consider this as a limitation of the evaluation process employed. We observed that the models which achieved an F1 score going from 0.50 to 0.60 still provide very high-quality outputs, perfectly comprehensible and in line with the instructions provided in the prompt. We started noticing a drop in the answers' quality in models with a F1 score of less than 0.50. The lower the score, the more the outputs were noisy (e.g., not respecting the requested format, or providing additional texts before and after the list of needs) or wrong (e.g., empty or containing wrong needs, usually duplicating other ones previously extracted or coming from the examples provided in the few-shot prompting).

### 4.2 Error Analysis

To gain deeper insights into the performance and limitations of each LLM, we conducted a manual error analysis focusing on the lowest 25% of outputs based on F1 (BERTScore) scores. This approach allowed us to examine, for each model, the outputs that fell into the first quartile of the F1 distribution, as they were likely to contain a higher concentration of impactful errors. We analysed these outputs by comparing the generated needs to the manually labelled ones and cross-referencing them with the original text from the TripAdvisor or Reddit posts where the models extracted the entities. We selected this subset because it was more representative of the problematic cases where the LLMs struggled the most. By manually reviewing these low-scoring outputs (with F1 ranging from 0 to 0.20), we aimed to identify the reasons behind their poor performance. We considered two aspects of the generated text: form and content.

For the form, we observed whether the generated texts respected the requested output format (a list of needs separated by comma, without additional text), because any additional and superfluous text could be a reason for a lower score, and which were the errors in the extracted needs.

For the content, we identified four categories of problems which could have caused a low score:

- Hallucination: instances where the model generated incorrect information, that was present neither in the text nor in the labels. These are often severe mistakes where the LLM introduces misleading or fabricated answers (Ji et al., 2023). An example is: given a post where the user asked about hotel recommendations, the model generated "*parking information*" (example taken from the output of Gemini few-shot).
- Procedural misinterpretation: situations where the model misunderstood the requested task, leading to responses that were completely unrelated to the input, such as asking to provide the input text (e.g., "*please provide me the text of your post*", from the output of Phi-2 few-shot), returning unrelated content (e.g., "*note: the text provided for each post is limited to one paragraph only and does not contain any information about their content or purpose beyond what has been described above*", from the output of Gemma CoT), or repeating the provided text instead of returning the extracted needs.
- Over-Specification: cases where the model's response, although not necessarily wrong, contained much more detailed or verbose information than required. While this is not a mistake per se, the disparity in specificity compared to the labels leads to lower evaluation scores, as the models' outputs deviate significantly from the ground truth. An example is: given the label "*daily activities advices*", the model generated "*recommendations for day trips activities Tokyo during warm weather*" (example taken from the output of Mistral few-shot).
- Different wording (evaluation mistake): in these instances, the model's output was correct, but used different wording from the label, leading to a lower score (due to the limitations of the automatic evaluation metrics). In these cases, the content aligned with the manually identified needs, but the variation in phrasing penalized the model in terms of similarity scores. For example, GPT-4 (few-shot) generated "*suggestions for trip planning*" for the label "*travel itinerary recommendations*".

For each examined instance, we assigned one of the four reasons of error and determined if the output format was correct. We then calculated the percentages of occurrences for each error type and format issue across the 25% lowest-scoring outputs for each model. This allowed us to quantify the frequency of content (Hallucinations, Over-Specification, Procedural misinterpretations, Different wording) and form (Wrong Output Format) problems, providing a clearer picture of the specific weaknesses of each model. Table 5 demonstrates the results of this analysis.

It is possible to observe that the model which achieved the highest scores in the evaluation phase (GPT-4, Mistral, GPT-3.5, and Gemini) also presents the highest rates of Over-Specification and Different wording, and low percentages of Hallucination (ranging from 5% to 25%), Procedural misinterpretations (ranging from 0% to 10%) and Wrong Output Format (0% for all of them). This suggests that while these models often add unnecessary or overly detailed content, which may reduce their automatic evaluation scores, they tend to provide correct information and to respect the output format requested.

Llama 2 13B, Llama 2 7B, Phi-2, and Mixtral, prompted through CoT, demonstrate high Hallucination and Procedural misinterpretations rates (ranging from 21% to 48% and from 6% to 35%, respectively). This means that, in the case of the 25% lowest-scoring outputs, these models frequently misunderstood the request or the input, or generated information not grounded in the input data. The same models, through few-shot prompting, have similar rates of Different wording (from 4% to 9%), but compared to their CoT counterparts, they have much higher percentages of Wrong Output Format (ranging from 9% to 49%), leading to lower evaluation scores. This seems to imply that the CoT technique was more effective in letting the models respect the requests of the prompt.

Gemma exhibits the highest rates of Wrong Output Format (74% for CoT and 69% for few-shot), indicating significant issues in following the requested style of answers.

| LLM | Prompt | Content problems | | | | Form problems |
|---|---|---|---|---|---|---|
| | | Hallucinat. | Procedural Misinterpret. | Over-Specification | Different wording | Wrong Output Format |
| GPT-4 | CoT | 17% | 0% | 42% | 41% | 0% |
| Mistral | CoT | 19% | 10% | 26% | 45% | 0% |
| Mistral | FS | 27% | 0% | 41% | 32% | 0% |
| Gemini | CoT | 25% | 9% | 54% | 12% | 0% |
| GPT-3.5 | FS | 6% | 0% | 79% | 15% | 0% |
| GPT-4 | FS | 5% | 7% | 80% | 8% | 0% |
| GPT-3.5 | CoT | 19% | 0% | 74% | 7% | 0% |
| Gemini | FS | 25% | 0% | 70% | 5% | 0% |
| Llama 2 7b | CoT | 45% | 8% | 33% | 14% | 0% |
| Phi-2 3b | CoT | 41% | 29% | 22% | 8% | 0% |
| Mixtral | CoT | 21% | 6% | 64% | 9% | 21% |
| Llama 2 13b | CoT | 48% | 35% | 11% | 6% | 28% |
| LLama 2 13b | FS | 26% | 5% | 62% | 7% | 9% |
| LLama 2 7b | FS | 16% | 15% | 64% | 5% | 49% |
| Mixtral | FS | 23% | 10% | 63% | 4% | 17% |
| Phi-2 3b | FS | 34% | 8% | 49% | 9% | 42% |
| Gemma7b | FS | 23% | 20% | 57% | 0% | 69% |
| Gemma7b | CoT | 22% | 67% | 11% | 0% | 74% |

Table 5. Breakdown of content and form problems across the 25% lowest-scoring outputs for each LLM and prompting technique. The table presents the percentage of instances falling into the four content problem categories, Hallucination, Over-Specification, Procedural Misinterpretation, Correct Answer (Evaluation Mistake), and the one of outputs that did not adhere to the correct format. *Each of the worst 25% model outputs was assigned to a content problem type only, therefore the sum of the percentages of these four sub-columns is 100%.

The findings underscore an alignment between manual error analysis and automatic evaluation, since the models which achieved the highest scores also present low percentages of content and form problems in their 25% worst outputs, while these issues are much more present in those which reached lower results, such as Gemma, Mixtral, and Llama. This error analysis also helps understanding the reasons of the performance of the LLMs, and the limitation of the metrics employed. In the case of the 25% worst outputs considered for each model, BERTScore, ROUGE, and BLEU were always very low. Through the error analysis, we could highlight that this could be for a good reason (Hallucination and Procedural misinterpretation), for a non-necessarily good reason (Wrong Output Format and Over-Specification, since the presence of additional and too specific information and a non-consistent format are the cause of the low scores, but they might not be a problem in some contexts of application), or even wrong (in the case of Different wording). Furthermore, the insights of the error analysis remark the importance of both model choice and prompting techniques in achieving high-quality outputs. While advanced models generally perform well, specific approaches such as CoT prompting and attention to output format are crucial for optimising response accuracy and consistency

### 4.3 Limitations

In considering the limitations of this study, it is important to recognize several key factors. Firstly, our hardware configuration has limited the performance of some models (Llama 2 13B and Mixtral). A more powerful computer infrastructure would have allowed us to test the models without any constraints and consider the possibility of fine-tuning them, to improve their performance. Secondly, the evaluation metrics such as BERTScore, ROUGE, and BLEU may not fully capture the subtle variations in how well the models perform, especially when it comes to understanding meaning and context. We believe that human evaluation remains crucial in assessing the quality and relevance of model-generated responses, particularly in specialised domains where additional context and specificity are essential. For this reason, we performed a manual error analysis, to understand the nature of the low-scoring outputs.

Furthermore, one of the key limitations of this study is the lack of direct comparison between LLMs and traditional non-LLM approaches for the task of customer needs extraction. Although LLMs have demonstrated strong potential in this area, we did not identify existing non-LLM methods that specifically address this task, particularly in the context of travel-related forums like TripAdvisor and Reddit. Developing a traditional approach for comparison would have broadened the research scope beyond our initial goals. Future work could explore this comparison to offer a more complete understanding of the relative advantages and limitations of both LLM and traditional techniques in customer needs analysis. However, we believe that the work of Pumpurs and Grabis (2024), which compared ChatGPT with traditional methods (e.g., the Kano model) in identifying user needs in digital-physical ecosystems, is applicable to our case. Their findings suggest that LLMs can enhance the efficiency of user needs extraction and categorization, supporting the applicability of LLMs for tasks similar to our own, even in different domains. This reinforces the relevance of our approach while acknowledging the importance of future benchmarking with traditional methods. Lastly, the effectiveness of the prompt optimisation technique employed may vary across the different LLMs, warranting further investigation into their impact on model performance and interpretability.

## 5. Conclusion

Our study demonstrates the potential of LLMs in automating language understanding and customer needs extraction processes, particularly in the domain of travels. Through the deployment of various LLMs, we have observed promising results for the task of extracting travel customer needs from TripAdvisor and Reddit posts. While closed models such as GPT-4, GPT-3.5, and Gemini exhibited robust performance in our analysis, open-source models like Mistral 7B, Llama 7B, and Phi-2 3B also showcased notable capabilities, despite their smaller sizes in terms of parameters. Particularly, Mistral 7B emerged as a standout performer, offering high-quality responses comparable to those of larger closed models while being deployable on more affordable computational infrastructure. Our study underscores the importance of selecting the most suitable LLM for specific business needs, considering factors such as performance, resource requirements, and the level of customization and security offered by open-source models. The findings presented offer insights for a diverse range of stakeholders. AI developers and NLP researchers can benefit from the evaluation of different language models, gaining insights into their strengths, limitations, and potential applications in customer needs extraction tasks. Market researchers and data analysts can utilize the methodologies and evaluation metrics outlined in the study to benchmark and assess the performance of language model-based solutions in similar tasks across various industries. Additionally, businesses and researchers operating in resource-constrained environments can identify suitable open-source models, such as Mistral 7B and Phi-2 3B, for implementing cost-effective AI solutions without compromising on performance. In conclusion, our work contributes to the growing body of research on LLMs and their applications in customer needs management. By providing empirical evidence of their effectiveness in extracting travel customer needs, we offer valuable insights for businesses seeking to leverage AI-driven solutions to enhance customer experience and optimize operational workflows.